\documentclass[runningheads]{llncs}
\usepackage[T1]{fontenc}
\usepackage{graphicx}
\usepackage{hyperref}
\hypersetup{
    colorlinks=true,
    linkcolor=black,
    urlcolor=black,
    citecolor=black
}

\def\equationautorefname~#1\null{Equation~(#1)\null}
\usepackage{url}
\usepackage{bm}
\usepackage{bbm}
\usepackage{xspace}
\usepackage{algorithmic}
\usepackage{algorithm}
\usepackage{mdwlist}    
\usepackage{paralist} 

\usepackage{adjustbox}
\usepackage{array}
\usepackage{booktabs}
\usepackage{multirow}

\usepackage{array,graphicx}
\usepackage{booktabs}
\usepackage{pifont}
\usepackage{color}

\usepackage{colortbl}
\definecolor{lightgray}{gray}{0.85}
\usepackage{multirow}

\usepackage{fancybox} 
\usepackage{amsmath}
\usepackage{amsfonts}

\usepackage{wrapfig}

\newcommand{\hide}[1]{}

\newcommand{\method}{\textsc{MoST}\xspace}

\newcommand{\cost}{CoST\xspace}
\newcommand{\tstovec}{TS2Vec\xspace}
\newcommand{\ssmf}{SSMF\xspace}
\newcommand{\informer}{Informer\xspace}
\newcommand{\last}{LaST\xspace}
\newcommand{\tlstm}{NET$^3$\xspace}
\newcommand{\tstcc}{TS-TCC\xspace}
\newcommand{\atd}{ATD\xspace}

\newcommand{\daily}{Daily\xspace}
\newcommand{\realdisp}{Realdisp\xspace}

\newcommand{\google}{Google Trends\xspace}
\newcommand{\us}{US\xspace}

\newcommand{\usdata}{US}
\newcommand{\worlddata}{World}
\newcommand{\ecommerce}{\usdata\#1\xspace}
\newcommand{\vod}{\usdata\#2\xspace}
\newcommand{\sweets}{\usdata\#3\xspace}
\newcommand{\facilities}{\usdata\#4\xspace}
\newcommand{\music}{\worlddata\#1\xspace}
\newcommand{\sns}{\worlddata\#2\xspace}
\newcommand{\apparel}{\worlddata\#3\xspace}

\newcommand{\nyc}{NYC-CB\xspace}
\newcommand{\air}{KnowAir\xspace}
\newcommand{\airone}{KnowAir\xspace}

\newcommand{\interloc}{M1D\xspace}
\newcommand{\interkey}{M2D\xspace}
\newcommand{\random}{Random\xspace}

\newcommand{\Chaind}{Channel-independence\xspace}

\newcommand{\Chadep}{Channel-dependence\xspace}

\newcommand{\romanone}{$\mathrm{(\hspace{.18em}i\hspace{.18em})}$\xspace}
\newcommand{\romantwo}{$\mathrm{(\hspace{.08em}ii\hspace{.08em})}$\xspace}
\newcommand{\romanthree}{$\mathrm{(i\hspace{-.08em}i\hspace{-.08em}i)}$\xspace}
\newcommand{\halfromanone}{$\mathrm{\hspace{.18em}i\hspace{.18em})}$\xspace}
\newcommand{\halfromantwo}{$\mathrm{\hspace{.08em}ii\hspace{.08em})}$\xspace}

\newcommand{\preprocess}{tensor slicing\xspace}
\newcommand{\Preprocess}{Tensor slicing\xspace}

\newcommand{\encoder}{slice feature encoder\xspace}
\newcommand{\Encoder}{Slice feature encoder\xspace}

\newcommand{\outputlayer}{aggregator\xspace}
\newcommand{\Outputlayer}{Aggregator\xspace}

\newcommand{\Mmode}{Multi-mode\xspace}

\newcommand{\norder}{$N^{th}$-order\xspace}
\newcommand{\triorder}{$3^{rd}$-order\xspace}
\newcommand{\tts}{tensor time series\xspace}
\newcommand{\TTS}{TTS\xspace}
\newcommand{\CL}{CL\xspace}

\newcommand{\MTS}{MTS\xspace}
\newcommand{\eq}[1]{Eq.~(#1)}
\newcommand{\tabl}[1]{Table~#1}
\newcommand{\fig}[1]{Fig.~#1}
\newcommand{\figA}[2]{Fig.~#1~(#2)}

\newcommand{\timestep}{T}

\newcommand{\fx}[1]{f(#1)}
\newcommand{\lossfunc}{\mathcal{L}}

\newcommand{\modelossloc}{\lossfunc^{(\dimloc)}_{M}}
\newcommand{\modelosskey}{\lossfunc^{(\dimkey)}_{M}}
\newcommand{\instanceloss}{\lossfunc_{I}}
\newcommand{\lossweight}{\alpha}

\newcommand{\tensor}{\mathcal{X}}
\newcommand{\unftensorloc}{\tensor^{(\dimloc)}}
\newcommand{\unftensorkey}{\tensor^{(\dimkey)}}

\newcommand{\tensorflat}{\matrx^{flat}}

\newcommand{\matrx}{X}
\newcommand{\featurevector}{V}
\newcommand{\featurevectorloc}{\featurevector^{(\dimloc)}}
\newcommand{\featurevectorkey}{\featurevector^{(\dimkey)}}
\newcommand{\featureveceach}{v}

\newcommand{\tememb}{W^{tem}}
\newcommand{\tokemb}{W^{(\dimloc)pro}}
\newcommand{\latentmatrix}{Z}

\newcommand{\tensorembloc}{\latentmatrix^{(\dimloc)emb}}

\newcommand{\tensorcauloc}{\latentmatrix^{(\dimloc)cau}}

\newcommand{\tensoravecauloc}{\latentmatrix^{(\dimloc)enc}}
\newcommand{\tensoravecaukey}{\latentmatrix^{(\dimkey)enc}}

\newcommand{\currenttimestep}{1}
\newcommand{\lookback}{w}
\newcommand{\predlength}{T_f}
\newcommand{\nontemporaldim}{d}
\newcommand{\dimloc}{\nontemporaldim_1}
\newcommand{\dimkey}{\nontemporaldim_2}
\newcommand{\dimNone}{\nontemporaldim_{N-1}}
\newcommand{\latentdim}{h}
\newcommand{\latentdimloc}{\latentdim_{\nontemporaldim}}
\newcommand{\latentdimkey}{\latentdim_{\nontemporaldim}}

\begin{document}
\title{
Disentangled Mode-Specific Representations for Tensor Time Series via Contrastive Learning
}
\titlerunning{Disentangled Mode-Specific Representations for Tensor Time Series}
%
\author{Kohei Obata\inst{1}\thanks{Corresponding author.} \and
Taichi Murayama\inst{2} \and
Zheng Chen\inst{1} \and
Yasuko Matsubara\inst{1} \and
Yasushi Sakurai\inst{1}}

\authorrunning{K. Obata et al.}
%
\institute{SANKEN, Osaka University, Osaka, Japan \\
\email{\{obata88,chenz,yasuko,yasushi\}@sanken.osaka-u.ac.jp} \and
Yokohama National University, Kanagawa, Japan \\
\email{murayama-taichi-bs@ynu.ac.jp}}
\maketitle              
\begin{abstract}

\Mmode \tts (\TTS) can be found in many domains,
such as search engines and environmental monitoring systems.
Learning representations of a \TTS benefits various applications, but it is also challenging since the complexities inherent in the tensor hinder the realization of rich representations.
In this paper, we propose a novel representation learning method designed specifically for \TTS, namely \method.
Specifically, \method uses a tensor slicing approach to reduce the complexity of the \TTS structure and learns representations that can be disentangled into individual non-temporal modes.
Each representation captures mode-specific features, which are the relationship between variables within the same mode, and mode-invariant features, which are in common in representations of different modes.
We employ a contrastive learning framework to learn parameters; the loss function comprises two parts intended to learn representation in a mode-specific way and mode-invariant way, effectively exploiting disentangled representations as augmentations.
Extensive experiments on real-world datasets show that \method consistently outperforms the state-of-the-art methods in terms of classification and forecasting accuracy.
Code is available at \url{https://github.com/KoheiObata/MoST}.
\end{abstract}
\section{Introduction}
    \label{010intro}
\begin{figure}[t]
    \centering
    \includegraphics[width=.8\linewidth]{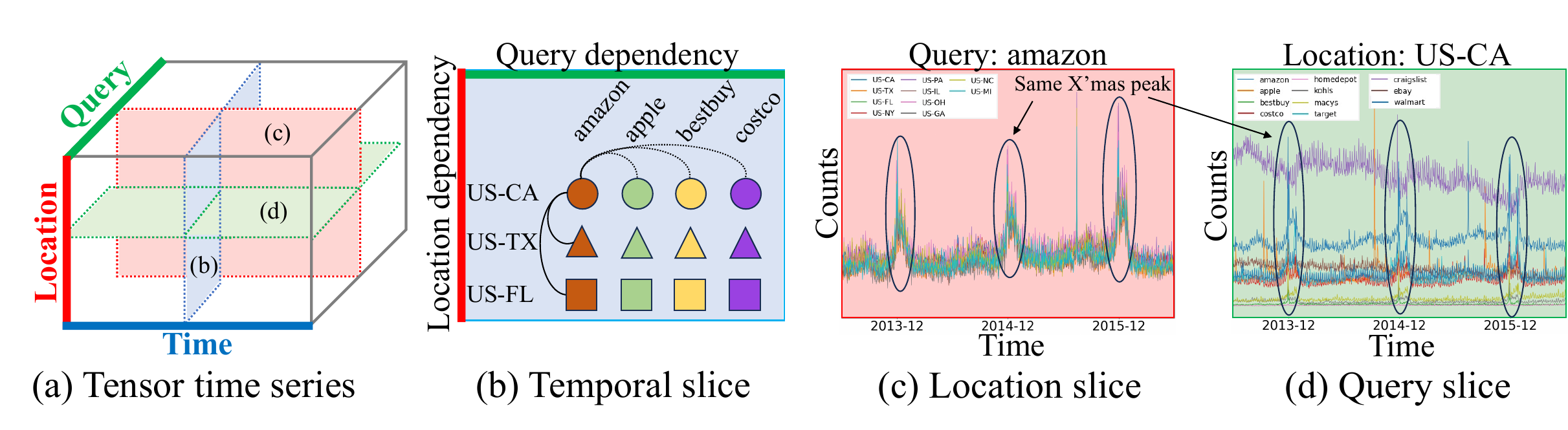}\\
    \vspace{-1em}
    \caption{Illustrations of a \tts and three slices along different modes.
    (a) A \tts with three modes: location, query, and time.
    (b) Location and query dependencies of a temporal slice. 
    (c), (d) Location and query slices have their own intra-mode dependencies, but temporal dependencies are common.
    }
    \label{fig:tensor}
    \vspace{-1em}
\end{figure}

    With the rapid growth of digital innovation for the collection of data, a time series sequence is generated from multiple attributes or modes, forming a tensor time series (\TTS).
\footnote{
Here, we mainly assume that the \TTS has three or more modes.
}
This complex data is increasingly prevalent across divers domains, ranging from environment monitoring~\cite{pm2.5} to financial analysis~\cite{financial}.
An instance of a \TTS is online activity data~\cite{compcube}, which records the search volumes of various queries at various locations for each timestamp, in the form $\{$Location, Query, Time$\}$ (\figA{\ref{fig:tensor}}{a}).
Each temporal slice of the \triorder \TTS (\figA{\ref{fig:tensor}}{b}) is a matrix where each non-temporal mode (i.e., location and query) has several variables.
As depicted in \figA{\ref{fig:tensor}}{c} and (d), each non-temporal mode slice consists of a multivariate time series (\MTS) that contains the interactions among corresponding variables and the dependencies between its past and future~\cite{missnet}.
Since \TTS has a multifaceted structure with intricate interactions, effectively learning meaningful representations from such data is a long-standing challenge.

Modeling \TTS requires us to consider the multiple non-temporal modes since tensors arise from the rich information of each mode.
Thus, tensor studies normally assume that different modes are associated with different types of dependencies~\cite{atd,facets,net3,dmm},
referred to as \textit{intra-mode dependencies}.
As shown in \figA{\ref{fig:tensor}}{b}, location and query dependencies are examples of intra-mode dependencies.
Moreover, while different mode slices in the \TTS have different intra-mode dependencies, there are common features invariant in modes as they originate from the same \TTS; that is, they have similar \textit{temporal dependencies}.
As shown in \figA{\ref{fig:tensor}}{c} and (d), the search volumes peak before and after every Christmas, meaning they have similar seasonality regardless of locations and queries.
Designing a framework that accommodates these intra-mode dependencies and temporal dependencies inherent in the \TTS is expected to facilitate the learning of rich representations.

Despite the number of studies in representation learning, existing methods take little account of the \TTS structure~\cite{timecontrastivereview}. 
Tensor decomposition is a conventional method for handling a tensor that can address intra-mode dependencies, which aims to provide a low-rank representation that minimizes a reconstruction error and additional penalty terms (e.g., Frobenius norm or KL-divergence)~\cite{hong2020generalized}.
Although a compact representation is an essential principle for feature reduction, such objective functions mainly aim for accurate reconstruction and are unsuitable for obtaining rich representation that can be applied to downstream tasks.
Therefore, recent works incorporate deep learning techniques to further facilitate the representation learning from tensor decomposition. 
For instance, \atd~\cite{atd}, a self-supervised representation learning method for tensors, combines tensor decomposition and contrastive learning (\CL) and succeeds in learning representations for the downstream classification task.
However, such tensor decomposition methods usually treat all modes equally, resulting in overlooking temporal dependencies. 
Thus, there is a need to develop a new approach to \TTS representation learning that effectively addresses the \TTS structure.

In this paper,
we present \textbf{\method},
which learns \textit{disentangled \textbf{Mo}de-\textbf{S}pecific representations for \textbf{T}ensor time series}.
Firstly, to reduce the complexity of the \TTS structure, 
we utilize a tensor slicing approach that slices a \TTS into each non-temporal mode to make sets of mode slices, where each set eventually converts into corresponding mode-specific representations.
After the slicing, slices are independently mapped into latent spaces via a \encoder so that the model can capture intra-mode dependencies,
and then are aggregated to generate mode-specific representations.
The parameters of the model are learned via a \CL framework, where we effectively exploit disentangled representations as augmentations to learn mode-specific and mode-invariant features inherent in the \TTS.
The contributions of this paper can be summarized as follows.
\begin{itemize}
\item We propose \method, a new representation learning method specifically designed to leverage the \TTS structure.
To the best of our knowledge, this is the first work to provide representations of \TTS via \CL.

\item To learn mode-specific representations, we introduce tensor slicing and propose corresponding contrastive losses.

\item We conduct extensive experiments on 11 real-world datasets to verify the effectiveness and efficiency of our proposed model in downstream tasks.
\end{itemize}

\section{Related work}
    \label{020related}
    \subsubsection{Tensor time series analysis}
Analyzing \TTS benefits a variety of applications, such as environmental monitoring~\cite{pm2.5,gmrl}, biomedical analysis~\cite{biotensor}, and online activity data~\cite{fluxcube}.
%
Traditional approaches to tensor analysis typically involve tensor/matrix decomposition: Tucker/CP decomposition and SVD~\cite{sofia,lowranktensor}.
They obtain a lower-dimensional representation that captures a good summarization of a tensor/matrix, but they do not consider the temporal dependencies~\cite{atd}.
To complement the temporal information, tensor/matrix decomposition is further combined with a dynamical system~\cite{mlds,facets,smf}.
For example, SSMF~\cite{ssmf} is an online forecasting method that combines a dynamical system with non-negative matrix factorization to capture seasonal trends in \TTS.
Recent work integrates deep learning models with tensor decomposition and is intended to learn invariant tensor features for forecasting~\cite{net3}.
However, since the dynamical system conveys the information recurrently from the one previous timestamp, they do not sufficiently capture long-term temporal dependencies.
%
Some models are designed to analyze a tensor from a specific modality by utilizing their intrinsic characteristics.
These include web data models that utilize the Lotka-Volterra equalitons~\cite{fluxcube,compcube}
and spatio-temporal models that utilize a pre-defined graph~\cite{spatiosurvey}.
Despite their success, these studies are data and task-specific; how to construct a general-purpose model that can extend to various \TTS remains a challenge.

\subsubsection{Time series contrastive learning}
Self-supervised learning enables transferable representations for diverse downstream tasks~\cite{timecontrastivereview}.
Inspired by the success in computer vision and NLP~\cite{stor,nlpreview}, \CL has recently emerged as a prominent technique for learning invariant representations in time series data~\cite{soft,mf-clr}.
For example, CPC~\cite{cpc}, a notable \CL method, learns representations by predicting future sequences in latent space,
and TNC~\cite{tnc} captures temporal smoothness by defining the neighborhood distribution.
However, they generate positive samples from neighboring timestamps and fail to capture global features in time series.
A recent successful method is TS2Vec~\cite{ts2vec}, which makes it possible to learn global features for time series by introducing hierarchical and instance losses.
The learned representations are generic to downstream tasks, but some methods focus particularly on learning task-specific representations, such as classification~\cite{mhccl,ts-tcc}, imputation~\cite{tider}, and change point detection~\cite{ts-cp2}. 
%
There is a recent promising trend that learns disentangled representations of time series through \CL.
As time series can be decomposed into seasonal and trend or spectral and temporal domains, CoST~\cite{cost} learns disentangled seasonal-trend representations for forecasting,
and TF-C~\cite{tf-c} and BTSF~\cite{btsf} exploit representations from spectral and temporal domains, to obtain global features in time series.

Our work distinguishes itself from these methods in the following aspects:
\halfromanone \method learns representations for \TTS that can be applied to any downstream tasks.
\halfromantwo \method learns disentangled representations capturing different \textit{mode-specific features} through a tensor slicing approach and contrasts the disentangled representations to learn \textit{mode-invariant features}.

\section{Proposed \method}
    \label{040model}
    \begin{figure}[t]
    \centering
    \includegraphics[width=.8\linewidth]{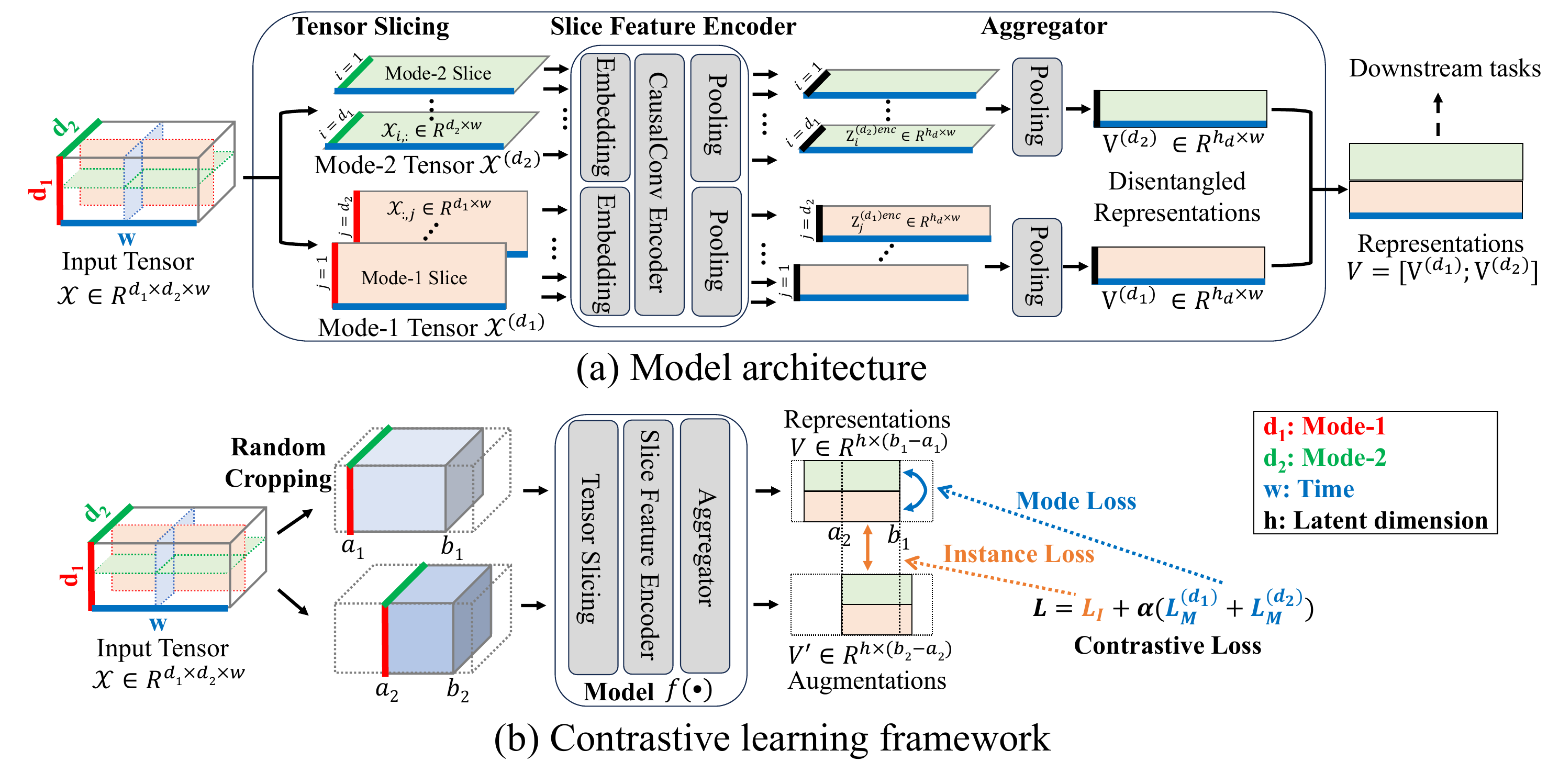}
    \vspace{-1.5em}
    \caption{
    (a) \method slices the \TTS along each non-temporal mode and independently feeds a slice into the \encoder to learn a representation of the slice. 
    Then, the \outputlayer is applied to summarize mode information.
    (b) The parameter of the model is learned via contrastive loss, which is composed of mode loss and instance loss.
    Mode loss utilizes the representations from different sliced tensors, while instance loss utilizes the representations generated by random cropping as contrastive augmentations.
    }
    \label{fig:architecture}
    \vspace{-1em}
\end{figure}

In this section,
we formally define the \TTS representation learning problem and then introduce our proposed approach, \method.
\begin{definition}[Tensor Time Series]
    A \TTS is an \norder tensor $\tensor \in \mathbb{R}^{\dimloc\times \dots \times\dimNone\times\timestep}$,
    where the $N^{th}$ mode is the time and its dimension is $\timestep$.
\end{definition}

\subsection{Problem formulation}
We consider a \triorder \TTS
$(\matrx_1, \ldots ,\matrx_\timestep) \in \mathbb{R}^{\dimloc\times\dimkey\times\timestep}$,
which consists of a set of observations
of $\dimloc$ and $\dimkey$ variables within the two non-temporal modes 
at $\timestep$ time points for the temporal mode.
\footnote{
In this work, we focus on a \triorder tensor for simplicity;
our method, however, can be generalized to an \norder tensor.
}
Given a \TTS with a $\lookback$ length window,
we aim to learn feature representations $\featurevector = \fx{\tensor}$, where
$\tensor = (\matrx_\currenttimestep, \ldots, \matrx_{\lookback}) \in \mathbb{R}^{\dimloc\times\dimkey\times\lookback}$,
$\featurevector = (\featureveceach_\currenttimestep, \ldots ,\featureveceach_{\lookback}) \in \mathbb{R}^{\latentdim\times\lookback}$,
and $\fx{\cdot}$ denotes the feature embedding function,
which projects $(\dimloc \times \dimkey)$-dimensional raw signals into an $\latentdim$-dimensional latent space for each time point.

Our \method aims to learn the disentangled mode-specific representations, i.e.,
$\featurevector = [\featurevectorloc ; \featurevectorkey] \in \mathbb{R}^{\latentdim \times \lookback}$,
where $\featurevectorloc \in \mathbb{R}^{\latentdimloc \times \lookback}$ is a mode-1-aware representation
and  $\featurevectorkey \in \mathbb{R}^{\latentdimkey \times \lookback}$ is a mode-2-aware representation,
such that $\latentdimloc = \latentdim / 2$.

\subsection{Model architecture} \label{ssec:architecture}
Our proposed model, illustrated in \figA{\ref{fig:architecture}}{a}, consists of three components:
\romanone \textbf{\Preprocess} transforms the tensor for learning disentangled mode-specific representations.
\romantwo \textbf{\Encoder} captures intra-mode and temporal dependencies and learns a representation of a slice individually.
\romanthree \textbf{\Outputlayer} summarizes the mode information to form a representation for a particular mode.

\subsubsection{\Preprocess} \label{ssec:slice}
We describe a method that effectively leverages the tensor structure, which helps learn rich representations from a \TTS.
Concretely, the tensor is processed into a set of mode-1,2 tensors, each of which represents a set of mode-1,2 slices.
The \preprocess is written as follows:
\begin{align}
    \{ \unftensorloc, \unftensorkey \} = \mathbf{Slice}(\tensor) , \label{eq:slice}
\end{align}
where $\unftensorloc$, is a mode-1 tensor,
and $\unftensorkey$ is a mode-2 tensor.
$\unftensorloc$ is a set of mode-1 slices, i.e., $\unftensorloc = \{ \tensor_{:,j} \}_{j=1}^{\dimkey}$, where $\tensor_{:,j} \in \mathbb{R}^{\dimloc \times \lookback}$ is a mode-1 slice at the $j$-th mode-2 of length $\lookback$, s.t., $j = 1, \dots, \dimkey$ is an index of slices.
Likewise, 
$\unftensorkey$ is a set of mode-2 slices, i.e., $\unftensorkey = \{ \tensor_{i,:} \}_{i=1}^{\dimloc}$, where $\tensor_{i,:} \in \mathbb{R}^{\dimkey \times \lookback}$ is a mode-2 slice at the $i$-th mode-1, s.t., $i = 1, \dots, \dimloc$.

\subsubsection{\Encoder} \label{ssec:encoder}
After slicing the \TTS, each mode-1,2 slice is fed independently into a \encoder, which we call a mode-independence (MI) approach.
Our \encoder maps the observed signals to the latent representations and learns intra-mode and temporal dependencies.
It consists of an embedding layer and a causal convolutional encoder.
Given that the procedure is identical for mode-1,2 tensors, $\unftensorloc$ and $\unftensorkey$, we only describe the procedure for mode-1 tensor, $\unftensorloc$.

Each mode-1 slice $\tensor_{:,j}$ of $\unftensorloc$ is mapped to the latent space of dimension $\latentdim$
via a trainable linear projection $\tokemb \in \mathbb{R}^{\latentdim \times \dimloc}$.
Then, a determinative additive temporal embedding~\cite{informer} $\tememb \in \mathbb{R}^{\latentdim \times \lookback}$ is used to monitor the temporal order of a series:
$\tensorembloc_{j} = \tokemb \tensor_{:,j} + \tememb$,
where $\tensorembloc_{j} \in \mathbb{R}^{\latentdim \times \lookback}$ denotes the input to be fed into a causal convolutional encoder.

To supply long-range temporal information,
we employ a causal convolutional encoder and capture information from different time scales~\cite{cost}.
The causal convolutional encoder consists of a stack of $L+1$ one-dimensional causal convolution blocks and an average-pooling layer.
We use $L=7$ in this paper.
Each block has $\latentdim$ input channels and $\latentdimloc$ output channels,
where the kernel size of the $k$-th block is $2^k$.
The parameters of the causal convolutional encoder are shared in different modes.
Then, an average-pooling layer is used to summarize the latent representation across different time scales.
\begin{align}
    \tensorcauloc_{j,k} = \mathbf{CausalConvBlock}(\tensorembloc_{j}, 2^k),  \label{eq:causalconv} \\
    \tensoravecauloc_{j} = \mathbf{AveragePooling}(\tensorcauloc_{j,0}, \dots ,\tensorcauloc_{j,L}). \label{eq:causalave}
\end{align}

\subsubsection{\Outputlayer} \label{ssec:output}
We have embedded representations of each slice $ \tensoravecauloc_{j} \in \mathbb{R}^{\latentdimloc \times \lookback}$, where $j = 1, \dots, \dimkey$ and,
$ \tensoravecaukey_{i} \in \mathbb{R}^{\latentdimkey \times \lookback}$, where $i = 1, \dots, \dimloc$,
respectively.
An average-pooling layer or a max-pooling layer is applied depending on the dataset to summarize the mode information.
Through this process,
we acquire mode-1-aware representations $\featurevectorloc$ and mode-2-aware representations $\featurevectorkey$.
\begin{align}
    \featurevectorloc = \mathbf{Pooling}(\tensoravecauloc_{1}, \dots ,\tensoravecauloc_{\dimkey}), \\
    \featurevectorkey = \mathbf{Pooling}(\tensoravecaukey_{1}, \dots ,\tensoravecaukey_{\dimloc}). \label{eq:feature}
\end{align}
We concatenate them and obtain the final representations
$\featurevector = [\featurevectorloc ; \featurevectorkey]$.

\subsection{Contrastive learning} \label{ssec:loss}
We utilize \CL, a self-supervised learning method, to optimize the entire network, resulting in more comprehensive representations of \TTS~\cite{simclr}.
We define a contrastive loss \eq{\ref{eq:loss}} that consists of \textbf{instance loss} \eq{\ref{eq:instanceloss}} and \textbf{mode loss} \eq{\ref{eq:modeloss}},
each of which is a variant of InfoNCE~\cite{cpc} that aims to capture the \textit{mode-specific features} and \textit{mode-invariant features}, respectively.

\subsubsection{Instance loss}
Instance loss aims to learn the mode-specific features.
As \figA{\ref{fig:architecture}}{b} shows, we apply random cropping to the input tensor before feeding it to the model and create two different augmentations.
Random cropping treats the representations at the same timestamp in two different augmentations as positive samples,
whereas it treats those of other time series samples at the same timestamp as negative samples~\cite{ts2vec}.
Concretely, we randomly sample two overlapping sequences from $\tensor$,
hence $(\matrx_{a_1}, \ldots, \matrx_{b_1})$,
$(\matrx_{a_2}, \ldots, \matrx_{b_2})$,
s.t. $0 < a_1 \leq a_2 \leq b_1 \leq b_2 \leq \lookback$.
The overlapped sequence $(\matrx_{a_2}, \ldots, \matrx_{b_1})$ is used as a positive sample to learn position-agnostic representations.
This is only applied in the training.

Let $\featurevector_{i,t}$ denote the representations for the $i$-th index of the input time series sample at the timestamp $t$.
Then, $\featurevector'_{i,t}$ is a positive sample of $\featurevector_{i,t}$,
which denotes the representations for the $i$-th sample at the same timestamp $t$ but from a different augmentation.
We calculate instance loss as:
\begin{align}
    \instanceloss = \frac{1}{BN} \sum_i^B \sum_t^N
    - log \frac{exp(\featurevector_{i,t} \cdot \featurevector'_{i,t})}
    {\sum_{j=1}^{B}(exp(\featurevector_{i,t} \cdot \featurevector'_{j,t}) + \mathbbm{1}_{[i \ne j]} exp(\featurevector_{i,t} \cdot \featurevector_{j,t}))} , \label{eq:instanceloss}
\end{align}
where $B$ is the batch size, $N = b_1 - a_2$ is the length of the overlapped sequence.
We use the representations of other samples at timestamp $t$ as negative samples.

\subsubsection{Mode loss}
To learn the mode-invariant features in different mode-specific representations, we treat representations of different modes at the same timestamp as a positive sample, while we treat those of other time series samples at the same timestamp of different modes as negative samples.
The approach attempts to minimize the similarity among different mode-specific representations of the same timestamp.

We describe the mode loss for mode-1 $\modelossloc$.
However, the mode loss for mode-2 $\modelosskey$ is defined as being the same as $\modelossloc$ but in a different mode.
Let $\featurevectorloc_{i,t}$ denote the mode-1-aware representations for the $i$-th index of the input time series sample at the timestamp $t$.
$\featurevectorkey_{i,t}$ is a positive sample of $\featurevectorloc_{i,t}$,
which denotes the mode-2-aware representations for the $i$-th sample at the same timestamp $t$.
The mode loss for mode-1 is defined as:
\begin{align}
    \modelossloc = \frac{1}{BN} \sum_i^B \sum_t^N
    - log \frac{exp(\featurevectorloc_{i,t} \cdot \featurevectorkey_{i,t})}
    {\sum_{j = 1}^{B} exp(\featurevectorloc_{i,t} \cdot \featurevectorkey_{j,t})} . \label{eq:modeloss}
\end{align}
We use the representations of other samples at timestamp $t$ from a different mode as negative samples.

\subsubsection{Contrastive loss}
The overall loss function is the weighted sum of the instance loss and the mode loss of each non-temporal mode.
\begin{align}
    \lossfunc = \instanceloss + \lossweight (\modelossloc + \modelosskey) , \label{eq:loss}
\end{align}
where $\lossweight$ is a hyper-parameter that controls the importance of the mode loss.

\section{Experiments}
    \label{050experiments}
    \subsection{Classification}
\subsubsection{Settings}
We follow \tstovec\cite{ts2vec}, where max-pooling is applied over all timestamps to obtain the representations for a sequence.
We then train a logistic regressor model with an L2 norm penalty.
%
We use two real-world \triorder \TTS motion sensor datasets.
(1) \textbf{\daily}
consists of five units and nine statistical features, with 19 activity labels.
\footnote{\url{https://archive.ics.uci.edu/dataset/256}}
(2) \textbf{\realdisp}
comprises nine units and 13 features, with 33 activity labels.
\footnote{\url{https://archive.ics.uci.edu/dataset/305}}
%
We compare our method with the following four state-of-the-art self-supervised representation learning methods for time series (\cost~\cite{cost}, \tstovec~\cite{ts2vec}, and \tstcc~\cite{ts-tcc}) and for tensor (\atd~\cite{atd}).
We use the classification accuracy (Acc) as an evaluation metric.

\begin{table}[t]
    \caption{
    Classification results with Acc.
    Best results are in \textbf{bold}, and second best results are \underline{underlined} (higher is better).
    The average of five runs is reported.
    }
    \label{table:classification}
    \centering
    \fontsize{6.5pt}{6.5pt}\selectfont
\begin{tabular}{l|lllll}
  \toprule
  Dataset & \method & \cost & \tstovec & \tstcc & \atd \\
  \midrule
  \daily & $\mathbf{0.726}$ & 0.164 & $\underline{0.688}$ & 0.478 & 0.249 \\
  \realdisp & $\mathbf{0.766}$  & 0.126 & $\underline{0.663}$ & 0.190 & 0.151 \\
  \bottomrule
\end{tabular}
\end{table}
\subsubsection{Results}
The evaluation results are shown in \tabl{\ref{table:classification}}.
In general, \method outperforms other baselines.
This is because \method exploits the \TTS structure, including intra-mode and temporal dependencies, resulting in learning rich representations.
\cost learns disentangled representations of trend and seasonality, but the representations are not suitable for classification.
\atd jointly learns tensor decomposition and representations through \CL, but fails to capture temporal information in \TTS.
\tstcc learns representations for time series classification and shows better results than \cost and \atd.
\tstovec is less accurate than \method because it does not exploit the \TTS structure.

\subsection{Forecasting}
\subsubsection{Settings}
We use the representations of the last timestamp to predict future $\predlength$ observations.
Then, following~\cite{ts2vec,cost}, we train a linear regression model with an L2 norm penalty to directly predict future observations. 
We use nine real-world datasets of \triorder \TTS.
(1) \textbf{\google} consists of daily web-search counts, including seven query sets from~\cite{fluxcube}, resulting in seven datasets: four from the 50 US states (\usdata\#1$\sim$\#3) and three from the top 50 countries by GDP (\worlddata\#1$\sim$\#4).
(2) \textbf{\air} consists of PM2.5 concentrations and weather forecasting data, 12 sensor values in total,
taken every three hours covering 184 cities~\cite{pm2.5}.
(3) \textbf{\nyc} contains NYC CitiBike\footnote{\url{https://s3.amazonaws.com/tripdata/index.html}}
ride trips obtained from the 25 starting and ending positions taken every three hours.
%
We compare our method with the following six state-of-the-art methods
for \CL (\cost~\cite{cost} and \tstovec~\cite{ts2vec}), end-to-end forecasting (\informer~\cite{informer} and \last~\cite{last}), and tensor/matrix decomposition (\tlstm~\cite{net3} and \ssmf~\cite{ssmf}).

\begin{table}[t]
    \caption{Forecasting performance (lower is better). 
    The average of five runs is reported.
    }
    \label{table:forecasting}
    \centering
    \fontsize{6.5pt}{6.5pt}\selectfont
  \begin{tabular}{ll|llllll|llll|llll}
  \toprule
  & \multirow{2}{*}{Methods} & \multicolumn{6}{c}{Self-supervised representation learning} & \multicolumn{4}{c}{End-to-end forecasting} & \multicolumn{4}{c}{Tensor forecasting} \\
  & & \multicolumn{2}{c}{\method} & \multicolumn{2}{c}{\cost} & \multicolumn{2}{c}{\tstovec} & \multicolumn{2}{c}{\informer} & \multicolumn{2}{c}{\last} & \multicolumn{2}{c}{\tlstm} & \multicolumn{2}{c}{\ssmf} \\
   & $\predlength$ &    MSE &    MAE &    MSE &    MAE &    MSE &    MAE &    MSE &    MAE &    MSE &    MAE &    MSE &    MAE &    MSE &    MAE \\
  \midrule
\multirow{3}{*}{\scalebox{0.7}{\rotatebox{90}{\ecommerce}}} & 2 week & $\mathbf{0.752}$ &	$\mathbf{0.590}$ &	$\underline{0.900}$ &	$\underline{0.668}$ &	1.153 &	0.806 &	1.364 &	0.818 &	1.003 &	0.694 &	1.501 &	0.915 &	1.380 &	0.824 \\
& 8 week & $\mathbf{0.811}$ &	$\mathbf{0.609}$ &	$\underline{1.002}$ & $\underline{0.685}$ &	1.133 &	0.751 &	1.125 &	0.735 &	1.060 &	0.731 &	2.143 &	1.058 &	1.353 &	0.815 \\
& 32 week & $\mathbf{0.949}$ &	$\mathbf{0.661}$ &	$\underline{1.132}$ &	$\underline{0.714}$ &	1.355 &	0.822 &	1.242 &	0.770 &	1.449 &	0.839 &	1.599 &	0.866 &	1.355 &	0.811 \\
\midrule
\multirow{3}{*}{\scalebox{0.7}{\rotatebox{90}{\vod}}} & 2 week & $\mathbf{0.653}$ &	$\mathbf{0.566}$ &	$\underline{0.676}$ &	$\underline{0.575}$ &	0.725 &	0.600 &	0.855 &	0.655 &	0.709 &	0.588 &	0.805 &	0.630 &	1.385 &	0.862 \\
& 8 week & $\mathbf{0.695}$ &	$\mathbf{0.580}$ &	0.729 & $\underline{0.595}$ &	0.766 &	0.613 &	0.891 &	0.670 &	$\underline{0.726}$ &	0.601 &	0.891 &	0.680 &	1.396 &	0.860 \\
& 32 week & $\mathbf{0.731}$ &	$\mathbf{0.597}$ &	$\underline{0.786}$ &	0.627 &	0.832 &	0.646 &	0.864 &	0.647 &	0.793 &	$\underline{0.621}$ &	1.013 &	0.733 &	1.562 &	0.913 \\
\midrule
\multirow{3}{*}{\scalebox{0.7}{\rotatebox{90}{\sweets}}} & 2 week & $\mathbf{0.591}$ &	$\mathbf{0.511}$ &	$\underline{0.628}$ &	$\underline{0.525}$ &	0.665 &	0.549 &	0.760 &	0.576 &	0.674 &	0.559 &	0.682 &	0.563 &	1.281 &	0.816 \\
& 8 week & $\underline{0.580}$ &	$\mathbf{0.506}$ &	0.610 & 0.519 &	0.625 &	0.530 &	0.597 &	0.535 &	$\mathbf{0.563}$ &	$\underline{0.516}$ &	0.697 &	0.576 &	1.268 &	0.815 \\
& 32 week & $\mathbf{0.521}$ &	$\mathbf{0.500}$ &	0.541 &	$\underline{0.509}$ &	0.563 &	0.524 &	0.578 &	0.548 &	$\underline{0.528}$ &	0.515 &	1.268 &	0.732 &	1.290 &	0.841 \\
\midrule
\multirow{3}{*}{\scalebox{0.7}{\rotatebox{90}{\facilities}}} & 2 week & $\mathbf{0.439}$ &	$\mathbf{0.500}$ &	$\underline{0.465}$ &	$\underline{0.517}$ &	0.497 &	0.541 &	0.723 &	0.656 &	0.517 &	0.555 &	0.547 &	0.567 &	1.841 &	0.998 \\
& 8 week & $\mathbf{0.448}$ &	$\mathbf{0.506}$ &	$\underline{0.477}$ & $\underline{0.526}$ &	0.515 &	0.552 &	0.683 &	0.630 &	0.527 &	0.553 &	0.647 &	0.625 &	1.954 &	1.030 \\
& 32 week & $\mathbf{0.463}$ &	$\mathbf{0.513}$ &	$\underline{0.486}$ &	$\underline{0.528}$ &	0.525 &	0.554 &	0.708 &	0.637 &	0.533 &	0.557 &	0.709 &	0.657 &	1.996 &	1.044 \\
\midrule
\multirow{3}{*}{\scalebox{0.7}{\rotatebox{90}{\music}}} & 2 week & $\mathbf{0.705}$ &	$\mathbf{0.402}$ &	$\underline{0.716}$ &	$\underline{0.420}$ &	0.796 &	0.470 &	0.761 &	0.457 &	0.802 &	0.464 &	$\underline{0.716}$ &	0.438 &	1.985 &	1.035 \\
& 8 week & $\mathbf{0.706}$ &	$\mathbf{0.418}$ &	$\underline{0.720}$ & $\underline{0.441}$ &	0.817 &	0.500 &	0.773 &	0.468 &	0.798 &	0.476 &	0.800 &	0.515 &	1.979 &	1.037 \\
& 32 week & $\mathbf{0.681}$ &	$\mathbf{0.435}$ &	0.734 &	0.484 &	0.877 &	0.548 &	$\underline{0.700}$ &	$\underline{0.446}$ &	0.822 &	0.508 &	0.966 &	0.681 &	1.764 &	0.968 \\
\midrule
\multirow{3}{*}{\scalebox{0.7}{\rotatebox{90}{\sns}}} & 2 week & $\underline{0.949}$ &	$\mathbf{0.573}$ &	0.953 &	0.591 &	1.172 &	0.687 &	1.129 &	0.652 &	$\mathbf{0.942}$ &	$\underline{0.584}$ &	1.449 &	0.792 &	1.663 &	0.848 \\
& 8 week & $\underline{0.982}$ &	$\mathbf{0.590}$ &	0.984 & $\underline{0.611}$ &	1.257 &	0.732 &	1.042 &	0.654 &	$\mathbf{0.951}$ &	0.632 &	1.505 &	0.822 &	1.608 &	0.836 \\
& 32 week & $\mathbf{1.104}$ &	$\mathbf{0.674}$ &	1.147 &	0.692 &	1.670 &	0.891 &	$\underline{1.124}$ &	$\underline{0.683}$ &	1.176 &	0.702 &	1.499 &	0.824 &	1.545 &	0.823 \\
\midrule
\multirow{3}{*}{\scalebox{0.7}{\rotatebox{90}{\apparel}}} & 2 week & $\mathbf{1.566}$ &	0.823 &	$\underline{1.593}$ &	$\mathbf{0.811}$ &	1.679 &	$\underline{0.818}$ &	1.670 &	0.844 &	1.634 &	0.838 &	1.599 &	$\underline{0.818}$ &	2.125 &	0.949 \\
& 8 week & $\underline{1.583}$ &	0.825 &	1.600 & $\mathbf{0.811}$ &	1.697 &	$\underline{0.819}$ &	$\mathbf{1.530}$ &	0.813 &	$\underline{1.583}$ &	$\underline{0.819}$ &	1.640 &	0.835 &	2.099 &	0.946 \\
& 32 week & $\underline{1.638}$ &	0.823 &	1.665 &	$\underline{0.814}$ &	1.765 &	0.833 &	$\mathbf{1.616}$ &	$\mathbf{0.813}$ &	1.752 &	0.857 &	1.678 &	0.830 &	2.116 &	0.949 \\
\midrule
\multirow{3}{*}{\scalebox{0.7}{\rotatebox{90}{\airone}}} & 2 week & $\mathbf{0.635}$ &	$\mathbf{0.537}$ &	$\underline{0.637}$ &	$\underline{0.552}$ &	0.873 &	0.703 &	0.819 &	0.667 &	0.710 &	0.595 &	1.024 &	0.796 &	7.006 &	2.163 \\
& 4 week & $\mathbf{0.651}$ &	$\mathbf{0.545}$ &	$\underline{0.660}$ & $\underline{0.564}$ &	0.881 &	0.706 &	0.861 &	0.698 &	0.725 &	0.608 &	1.007 &	0.785 &	7.042 &	2.170 \\
& 8 week & $\mathbf{0.659}$ &	$\mathbf{0.550}$ &	$\underline{0.677}$ &	$\underline{0.574}$ &	0.867 &	0.697 &	0.862 &	0.697 &	0.691 &	0.586 &	1.023 &	0.793 &	7.116 &	2.185 \\
\midrule
\multirow{3}{*}{\scalebox{0.7}{\rotatebox{90}{\nyc}}} & 1 day & $\mathbf{0.905}$ &	$\mathbf{0.485}$ &	$\underline{0.926}$ &	$\underline{0.489}$ &	1.172 &	0.556 &	1.507 &	0.722 &	0.989 &	0.545 &	1.355 &	0.663 &	1.817 &	0.652 \\
& 1 week & $\mathbf{0.950}$ &	$\mathbf{0.497}$ &	0.958 & $\underline{0.501}$ &	1.203 &	0.564 &	1.568 &	0.753 &	$\underline{0.957}$ &	0.539 &	1.310 &	0.639 &	1.844 &	0.660 \\
& 2 week & $\underline{0.981}$ &	$\mathbf{0.506}$ &	0.984 &	$\underline{0.510}$ &	1.228 &	0.572 &	1.506 &	0.742 &	$\mathbf{0.959}$ &	0.540 &	1.271 &	0.612 &	1.874 &	0.669 \\
\bottomrule
\end{tabular}

\end{table}
\subsubsection{Results}
The experimental results are presented in \tabl{\ref{table:forecasting}}.
Overall, our model outperforms most of the comparative methods,
confirming the benefits of our model architecture.
\cost and \tstovec, which are \CL methods for time series,
are less accurate than our method.
The most significant difference between our method and the others is that \method uses a \TTS structure that significantly contributes to improved accuracy.

\begin{table}[t]
    \caption{Performance of forecasting and classification for the different (a) tensor slicing and MI approaches, (b) architectures, and (c) loss functions.}
    \label{table:percent_archi}
    \centering
    \fontsize{6.5pt}{6.5pt}\selectfont
\begin{tabular}{ll|ll|l}
  \toprule
  && \multicolumn{2}{c}{\textit{Forecasting}} & \textit{Classification} \\
  && Avg. MSE & Avg. MAE & Avg. Acc \\
  \midrule
\multicolumn{2}{c}{\method} & \multicolumn{1}{c}{$\mathbf{0.636}$}	& \multicolumn{1}{c}{$\mathbf{0.553}$}	& \multicolumn{1}{c}{$\mathbf{0.746}$} \\
  \midrule
\multirow{5}{*}{(a)} & w/o Mode-2 dependency (\interloc) & 0.811 (-27.5\%)	& 0.630 (-13.9\%)	& 0.714 (-4.4\%)	 \\
& w/o Mode-1 dependency (\interkey) & 0.650 (-2.2\%)	& 0.560 (-1.1\%)	& 0.599 (-19.8\%)	 \\
& \random & 0.656 (-3.2\%)	& 0.556 (-2.3\%)	& 0.626 (-16.2\%)	 \\
& \Chaind (CI) & 0.651 (-2.3\%)	& 0.559 (-1.0\%)	& 0.667 (-10.6\%)	 \\
& \Chadep (CD) & 0.912 (-43.3\%)	& 0.666 (-20.0\%)	& 0.405 (-45.7\%)	 \\
  \midrule
\multirow{2}{*}{(b)} & w/o Temporal Embedding & 0.693 (-8.9\%)	& 0.561 (-4.4\%)	& -	 \\
& w/o Causal Conv Encoder & 0.726 (-14.1\%)	& 0.593 (-8.6\%)	& 0.536 (-28.1\%)	 \\
  \midrule
\multirow{4}{*}{(c)} & w/o Instance Loss & 0.644 (-1.2\%)	& 0.557 (-0.7\%)	& 0.730 (-2.2\%)	 \\
& w/o Mode Loss & 0.636 (-0.0\%)	& 0.553 (-0.0\%)	& 0.732 (-1.9\%)	 \\
& MSE Loss & 0.823 (-29.1\%)	& 0.643 (-16.2\%)	& -	 \\
& MSE and Contrastive Loss & 1.342 (-111.0\%)	& 0.852 (-54.2\%)	& -	 \\
  \bottomrule
\end{tabular}

\end{table}
\subsection{Ablation study} \label{ssec:ablation}
We describe ablation studies concerning the architecture of our method.
\tabl{\ref{table:percent_archi}} presents the mean results of the classification datasets and the \us datasets on all forecast horizon settings with different architectures.

We first perform an ablation study to understand the effects of the tensor slicing and MI approaches.
We conduct a comparison with the five alternative preprocessing methods.
\textbf{Mode-1/2 dependency (M1D/M2D)}: Feeds only the mode-1/2 tensor $\unftensorloc/\unftensorkey$ into the \encoder.
\textbf{\random}: Creates a randomly organized tensor by flattening, randomizing variables, and restoring the original shape, applied once before learning.
\textbf{\Chaind (CI)}: Feeds $\{\tensor_{1,1},\tensor_{1,2}, \dots ,\tensor_{\dimloc,\dimkey} \}$ independently into the \encoder.
\textbf{\Chadep (CD)}: Feeds a flattened tensor $\tensorflat \in \mathbb{R}^{\dimloc\dimkey\times\lookback}$ into the \encoder.
\tabl{\ref{table:percent_archi}}(a) demonstrates the significant impact of our approaches on improving prediction and classification accuracy.
Comparison with \interloc and \interkey underscores the importance of both mode-1 and mode-2 dependencies for rich representations.
Additionally, comparisons with \random, CI, and CD demonstrate the superiority of our MI approach, which leverages the \TTS structure to learn intra-mode dependencies effectively.

We evaluate the impact of our architecture and loss functions.
\tabl{\ref{table:percent_archi}}(b) reveals that both temporal embedding and the causal convolutional encoder enhance performance.
\tabl{\ref{table:percent_archi}}(c) presents results w/o instance and mode losses and end-to-end training using forecasting loss functions.
Instance loss proves crucial for forecasting, while both losses improve classification accuracy.
Furthermore, our model, which optimizes representations and forecasting separately, outperforms end-to-end training with supervised MSE forecasting loss.

\subsection{Case study}
We use synthetic data to compare the representation ability of \method, \cost and \tstovec to learn intra-mode dependencies 
We generate a synthetic dataset by defining three different intra-mode dependencies for each non-temporal mode, and taking the cross product to form nine time series.

\begin{figure}[t]
    \centering
    \includegraphics[width=.45\linewidth]{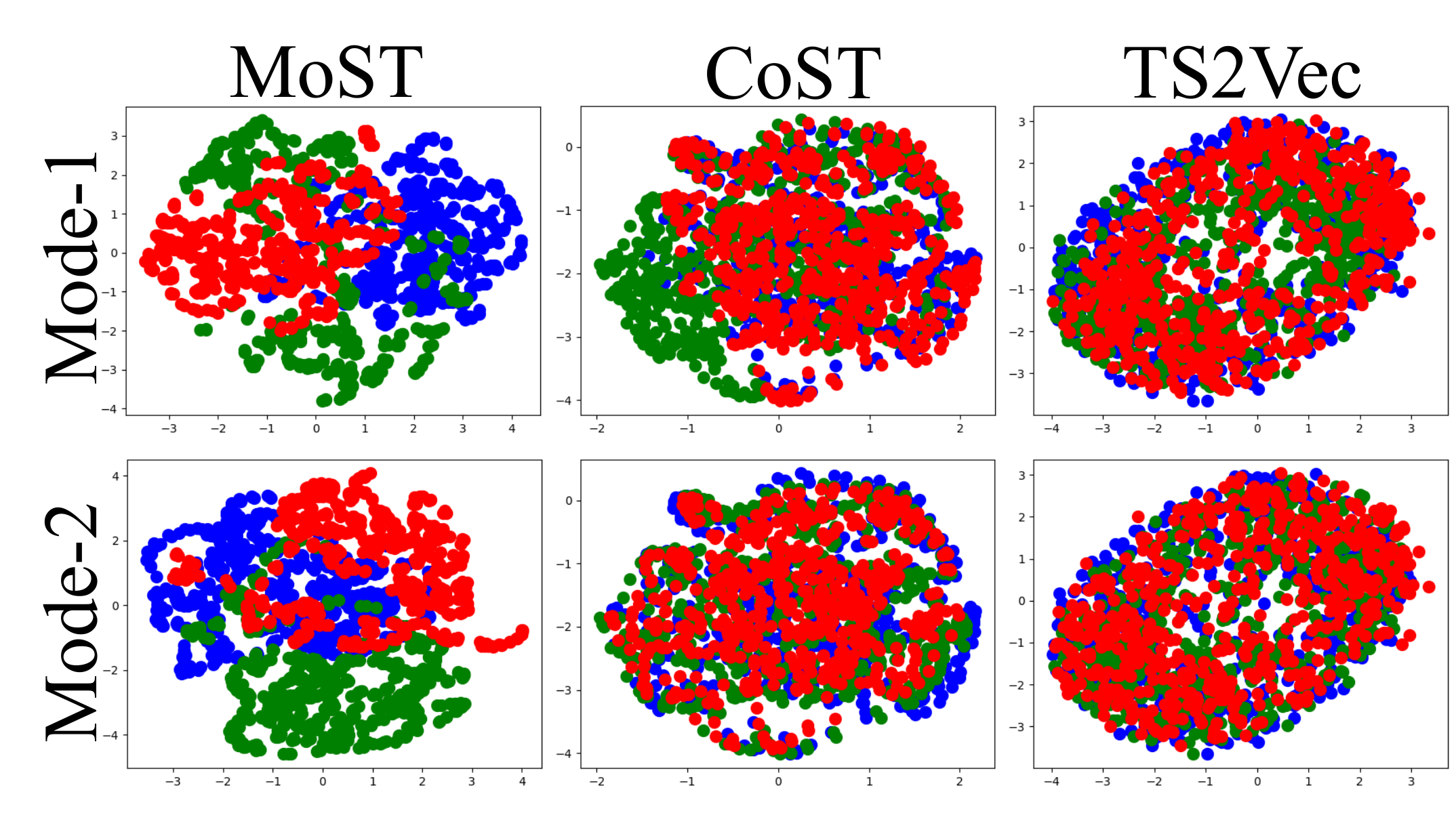}
    \vspace{-1em}
    \caption{
    t-SNE visualizations of the learned representations.
    The colors represent the three distinct intra-mode dependencies for (top) mode-1 and (bottom) mode-2.
    For \method, representations of mode-1 $\featurevectorloc$ and mode-2 $\featurevectorkey$ are shown. 
    }
    \label{fig:visual}
\end{figure}

We visualize the learned representations with t-SNE~\cite{t-sne} in \fig{\ref{fig:visual}}.
\cost and \tstovec are unable to distinguish between different intra-mode dependencies.
In contrast, our approach is able to discriminate between various intra-mode dependencies in the latent space thanks to its disentangled representations.

\section{Conclusion}
    \label{060conclusions}
    In this study,
we proposed \method, a representation learning method that learns disentangled mode-specific representations for \TTS.
\method consists of the tensor slicing approach and a newly defined contrastive loss to effectively model the \TTS.
Thanks to our proposed architecture, \method captures intra-mode and temporal dependencies.
The learned representations can be applied to any downstream tasks, including classification and forecasting,
and extensive experiments show that our method outperforms the state-of-the-art methods.
\method has the potential to contribute to enhancing analysis and our understanding of \TTS.

{\footnotesize
\subsubsection{Acknowledgments.}
The authors would like to thank the anonymous referees for their valuable comments and helpful suggestions.
This work was supported by
JSPS KAKENHI Grant-in-Aid for Scientific Research Number
JP23K16889, 
JP24K20778, 
JST CREST JPMJCR23M3.   
}

%
%
%
\bibliographystyle{splncs04}
\bibliography{ref_contrastive}

@inproceedings{pm2.5,
  title={Pm2. 5-gnn: A domain knowledge enhanced graph neural network for pm2. 5 forecasting},
  author={Wang, Shuo and others},
  booktitle={SIGSPATIAL},
  pages={163--166},
  year={2020}
}

@article{ssmf,
  title={Ssmf: Shifting seasonal matrix factorization},
  author={Kawabata, Koki and others},
  journal={NeurIPS},
  volume={34},
  pages={3863--3873},
  year={2021}
}

@inproceedings{smf,
  title={Smf: Drift-aware matrix factorization with seasonal patterns},
  author={Hooi, Bryan and others},
  booktitle={SIAM International Conference on Data Mining},
  pages={621--629},
  year={2019},
}

@inproceedings{facets,
  title={Facets: Fast comprehensive mining of coevolving high-order time series},
  author={Cai, Yongjie and others},
  booktitle={KDD},
  pages={79--88},
  year={2015}
}

@article{mlds,
  title={Multilinear dynamical systems for tensor time series},
  author={Rogers, Mark and others},
  journal={NeurIPS},
  volume={26},
  year={2013}
}

@inproceedings{last,
  title={Learning Latent Seasonal-Trend Representations for Time Series Forecasting},
  author={Wang, Zhiyuan and others},
  booktitle={NeurIPS},
  year={2022}
}

@inproceedings{cost,
    title={Co{ST}: Contrastive Learning of Disentangled Seasonal-Trend Representations for Time Series Forecasting},
    author={Gerald Woo and others},
    booktitle={ICLR},
    year={2022},
}

@inproceedings{ts2vec,
  title={Ts2vec: Towards universal representation of time series},
  author={Yue, Zhihan and others},
  booktitle={AAAI},
  pages={8980--8987},
  year={2022}
}

@inproceedings{btsf,
  title={Unsupervised time-series representation learning with iterative bilinear temporal-spectral fusion},
  author={Yang, Ling and others},
  booktitle={ICML},
  pages={25038--25054},
  year={2022},
  organization={PMLR}
}

@inproceedings{mhccl,
  author       = {Qianwen Meng and others},
  title        = {{MHCCL:} Masked Hierarchical Cluster-Wise Contrastive Learning for
                  Multivariate Time Series},
  booktitle    = {{AAAI}},
  pages        = {9153--9161},
  year         = {2023}
}

@inproceedings{stor,
  title={Contrastive spatio-temporal pretext learning for self-supervised video representation},
  author={Zhang, Yujia and others},
  booktitle={AAAI},
  pages={3380--3389},
  year={2022}
}

@inproceedings{
tider,
title={Multivariate Time-series Imputation with Disentangled Temporal Representations},
author={SHUAI LIU and others},
booktitle={ICLR},
year={2023},
}

@inproceedings{
tf-c,
title={Self-Supervised Contrastive Pre-Training For Time Series via Time-Frequency Consistency},
author={Xiang Zhang and others},
booktitle={NeurIPS},
year={2022},
}

@inproceedings{
ts-cp2,
title={Time Series Change Point Detection with Self-Supervised Contrastive Predictive Coding}, 
author={Deldari, Shohreh and others},
year = {2021},
booktitle = {WWW},
numpages = {12},
}

@inproceedings{simclr,
  title={A simple framework for contrastive learning of visual representations},
  author={Chen, Ting and others},
  booktitle={ICML},
  pages={1597--1607},
  year={2020},
  organization={PMLR}
}

@inproceedings{informer,
  author    = {Haoyi Zhou and others},
  title     = {Informer: Beyond Efficient Transformer for Long Sequence Time-Series Forecasting},
  booktitle = {AAAI},
  pages     = {11106--11115},
  year      = {2021},
}

@article{lowranktensor,
  title={Fast multivariate spatio-temporal analysis via low rank tensor learning},
  author={Bahadori, Mohammad Taha and others},
  journal={NeurIPS},
  volume={27},
  year={2014}
}

@article{t-sne,
  title={Visualizing data using t-SNE.},
  author={Van der Maaten, Laurens and others},
  journal={Journal of machine learning research},
  volume={9},
  number={11},
  year={2008}
}

@inproceedings{sofia,
  title={Robust factorization of real-world tensor streams with patterns, missing values, and outliers},
  author={Lee, Dongjin and others},
  booktitle={ICDE},
  pages={840--851},
  year={2021},
  organization={IEEE}
}

@inproceedings{gmrl,
  author       = {Jiewen Deng and others},
  title        = {Learning Gaussian Mixture Representations for Tensor Time Series Forecasting},
  booktitle    = {IJCAI},
  pages        = {2077--2085},
  year         = {2023},
}

@article{atd,
  title={ATD: Augmenting CP Tensor Decomposition by Self Supervision},
  author={Yang, Chaoqi and others},
  journal={NeurIPS},
  volume={35},
  pages={32039--32052},
  year={2022}
}

@article{hong2020generalized,
  title={Generalized canonical polyadic tensor decomposition},
  author={Hong, David and others},
  journal={SIAM Review},
  volume={62},
  number={1},
  pages={133--163},
  year={2020},
  publisher={SIAM}
}

@article{biotensor,
  title={Image analysis and machine learning in digital pathology: Challenges and opportunities},
  author={Madabhushi, Anant and others},
  journal={Medical image analysis},
  volume={33},
  pages={170--175},
  year={2016},
  publisher={Elsevier}
}

@article{timecontrastivereview,
  title={Self-Supervised Learning for Time Series Analysis: Taxonomy, Progress, and Prospects},
  author={Zhang, Kexin and others},
  journal={arXiv preprint arXiv:2306.10125},
  year={2023}
}

@article{nlpreview,
  title={A Primer on Contrastive Pretraining in Language Processing: Methods, Lessons Learned, and Perspectives},
  author={Rethmeier, Nils and others},
  journal={ACM Computing Surveys},
  volume={55},
  number={10},
  pages={1--17},
  year={2023},
  publisher={ACM New York, NY}
}

@article{spatiosurvey,
  title={Deep learning for spatio-temporal data mining: A survey},
  author={Wang, Senzhang and others},
  journal={TKDE},
  year={2020},
  publisher={IEEE}
}

@article{financial,
  title={Financial time series},
  author={Tsay, Ruey S},
  journal={Encyclopedia of Statistical Sciences},
  volume={4},
  year={2004},
  publisher={Wiley Online Library}
}

@inproceedings{net3,
  author       = {Baoyu Jing and others},
  title        = {Network of Tensor Time Series},
  booktitle    = {WWW},
  pages        = {2425--2437},
  year         = {2021},
}

@inproceedings{fluxcube,
  author       = {Taichi Murayama and others},
  title        = {Mining Reaction and Diffusion Dynamics in Social Activities},
  booktitle    = {CIKM},
  pages        = {1521--1531},
  year         = {2022},
}

@inproceedings{tnc,
  author       = {Sana Tonekaboni and others},
  title        = {Unsupervised Representation Learning for Time Series with Temporal
                  Neighborhood Coding},
  booktitle    = {ICLR},
  year         = {2021},
}

@inproceedings{ts-tcc,
  author       = {Emadeldeen Eldele and others},
  title     = {Time-Series Representation Learning via Temporal and Contextual Contrasting},
  booktitle = {IJCAI},
  pages     = {2352--2359},
  year      = {2021},
}

@article{cpc,
  author       = {A{\"{a}}ron van den Oord and others},
  title        = {Representation Learning with Contrastive Predictive Coding},
  journal      = {CoRR},
  volume       = {abs/1807.03748},
  year         = {2018},
  url          = {http://arxiv.org/abs/1807.03748},
  eprinttype    = {arXiv},
  eprint       = {1807.03748},
}

@inproceedings{compcube,
author = {Matsubara, Yasuko and others},
title = {Non-Linear Mining of Competing Local Activities},
year = {2016},
booktitle = {WWW},
pages = {737–747},
numpages = {11},
}

@inproceedings{soft,
title={Soft Contrastive Learning for Time Series},
author={Seunghan Lee and others},
booktitle={ICLR},
year={2024},
}

@InProceedings{mf-clr,
  title = 	 {{MF}-{CLR}: Multi-Frequency Contrastive Learning Representation for Time Series},
  author =       {Duan, Jufang and others},
  booktitle = 	 {ICML},
  pages = 	 {11918--11939},
  year = 	 {2024},
  volume = 	 {235},
  series = 	 {Proceedings of Machine Learning Research},
  month = 	 {21--27 Jul},
  publisher =    {PMLR},
}

@inproceedings{dmm,
  title={Dynamic Multi-Network Mining of Tensor Time Series},
  author={Obata, Kohei and others},
  booktitle={WWW},
  pages={4117--4127},
  year={2024}
}

@inproceedings{missnet,
  title={Mining of switching sparse networks for missing value imputation in multivariate time series},
  author={Obata, Kohei and others},
  booktitle={KDD},
  pages={2296--2306},
  year={2024}
}

\end{document}